\documentclass{article}
\usepackage{graphicx, amsmath, algorithm, tikz, url}
\usepackage[noEnd=true, indLines=true]{algpseudocodex}
\usepackage[margin=25mm]{geometry}

\newcommand{\nout}{n_{\textnormal{out}}} \renewcommand{\L}{\operatorname{L}} \newcommand{\dd}{\mathrm d}
\newcommand{\dzideta}{\frac {\dd z_{\alpha i}} {\dd \eta}} \newcommand{\dzjdeta}{\frac {\dd z_{\alpha j}} {\dd \eta}} \newcommand{\dqdeta}{\frac {\dd q_\alpha} {\dd \eta}}
\newcommand{\bignorm}[1]{\Big\lVert#1\Big\rVert} \newcommand{\tst}{{\textnormal{test}}}
\newcommand{\etaso}{\hat \eta_{\text{so}}} \newcommand{\etacs}{\hat \eta_{\text{cs}}}
\DeclareMathOperator{\Hess}{H} \DeclareMathOperator{\Jac}{J} 

\title{Stepsize modulation for cross entropy from local quadratic optimization}

\begin{document}
\author{Bruno Bauwens\footnote{National Research University Higher School of Economics, Faculty of Computer Science, Moscow, Russia;
This work/article is an output of a research project (HSE-BR-2025-024) implemented as part of the Basic Research Program at HSE University.
This research was supported in part through computational resources of HPC facilities at HSE University.
}, Viacheslav Krivodub\footnote{Luxoft Serbia.}, and Bingyu Zhang}
\date{\today}
\maketitle


\begin{abstract}
  Two methods are presented that perturb a given learning rate schedule in stochastic gradient descent with cross entropy loss. 
  Both methods use an extra forward pass to estimate various derivatives of the minibatch loss. 

  The first method selects a value $\eta$ by linearising the softmax values and minimizing the squared error $\tfrac 1 2 ||y - q - \eta q'||_F^2$, 
  where $q$ is the matrix representing the softmax values of the minibatch and $y$ is the matrix with one-hot encodings of the labels. 
  The second method considers a quadratic Taylor approximation of the minibatch loss under the outer approximation of the Hessian. 
  Thus, it minimizes $L(0) + \eta L'(0) + \tfrac 1 2 \eta^2 L''(0)$. 
  Here, the derivatives $L'$ and $L''$ are computed from the derivatives of the logits (using the outer approximation of the Hessian of the loss). 
  We observe that both methods give very similar values of $\eta$. 

  Finally, these estimated $\eta$ are used to slightly increase or decrease the learning rate of an existing schedule. 
  The method is evaluated on Cifar10 and Cifar100 datasets for training Resnets, where they decrease the loss faster than the schedule that is perturbed and increase monotonicity. 
  For Cifar10 it also leads to a small improvement of test accuracy, but in Cifar100 test accuracy remains the same.\footnote{Code: \url{https://github.com/Brbauwens/adaptive_stepsize_modular}}
\end{abstract}

\section{Introduction}

The selection of the learning rate is a very consequential decision when training neural nets. 
Despite much research in automatically selecting step sizes, typical training with convolutional nets 
is done with some fixed schedule for which maximum learning rate and weight decays are tuned. 
In this paper, we ask whether one can modulate the standard cosine schedule using batch dependant stepsizes to improve training. 
More precisely, can we inspect certain properties of a minibatch and use it to decide whether a bigger or smaller stepsize should be used compared to the scheduled value of the epoch. 
Can this speed up training and increase test accuracy?

\subsection*{The Polyak stepsize}

Consider a loss function $L(\theta)$. 
We aim to find a value of $\theta$ that makes $L$ as small as possible. 
The gradient descent method starts from an arbitrary $\theta_0$ and iteratively computes
\[
   \theta_{t+1} = \theta_t - \eta_t \nabla L(\theta_t). 
\]
Already in 1969, Polyak proposed to automatically select the stepsize using a linear approximation of the loss function~\cite{Polyak1969minimization}. 
Consider the linear approximation
\[
  L(\theta_t - w) \approx L(\theta_t) - \nabla L(\theta_t) \cdot w.
\]
For $w = \eta_t \nabla L(\theta_t)$, this simplifies to
\[
  L(\theta_{t+1}) \approx L(\theta_t) - \eta_t ||\nabla L(\theta_t)||^2. 
\]
Solving for $L(\theta_{t+1}) \approx 0$, gives
\[
   \eta_t = \frac {L(\theta)} {||\nabla L(\theta_t)||^2}. 
\]
It was directly clear that if $L$ is smooth and strongly convex, then gradient descent with this stepsize converges exponentially fast in~$t$~\cite{Polyak1969minimization}. 
A bit later, Polyak proved that for all convex functions (possibly non-smooth) this method has $O(1/\sqrt{t})$ convergence: $\liminf_t \sqrt{t}(L(\theta_t) - L(\theta^*)) = 0$. 

In machine learning, the above gradients are estimated on small random selections of the training data, called minibatches. 
This means that for each step, a random approximation $L_t(\theta)$ of the loss is generated. 
The Polyak stepsize under this function is called the stochastic Polyak stepsize, for which $O(1/t)$ convergence was proven under star-convexity of the loss~\cite[theorem 4.4]{Gower2021sgd}, and this is locally satisfied in deep neural nets~\cite{Kleinberg2018alternative,Zhou2019SGD}. 

\begin{figure}
    \begin{center}
    \begin{tikzpicture}[domain=0:1.7,scale=1.5]
      \useasboundingbox (-1,-.2) rectangle (2,2); 
      \draw[very thin,color=gray] (0,0) grid [step=0.5] (1.7,1.8);

      \draw[->] (-.2,0) -- (1.8,0) node[right] {$\eta$};
      \draw[->] (0,-0.2) -- (0,1.9); 

      \draw[color=orange]    plot (\x,1.5 - \x)    node[right, yshift=-4pt] {$L(0) + \eta L'(0)$\vspace{-2em}};
      \draw[color=blue] plot (\x,{1.5 + 0.5 * \x * \x - \x)}) node[right] {$L(\eta)$};

      \draw[blue, dashed] (1,1) -- (1,0) node[below, black] {$\eta^*$};

      \node[circle, fill, inner sep=1pt, label={left:$L(0)$}] at (0, 1.5) {};
      \node[circle, fill, inner sep=1pt, blue] at (1, 1) {};
      \node[circle, fill, inner sep=1pt, orange] at (1.5, 0) {};
    \end{tikzpicture}
  \end{center}
  \caption{The Polyak stepsize over estimates on convex strictly positive functions}
  \label{fig:PolyakOverestimates}
\end{figure}
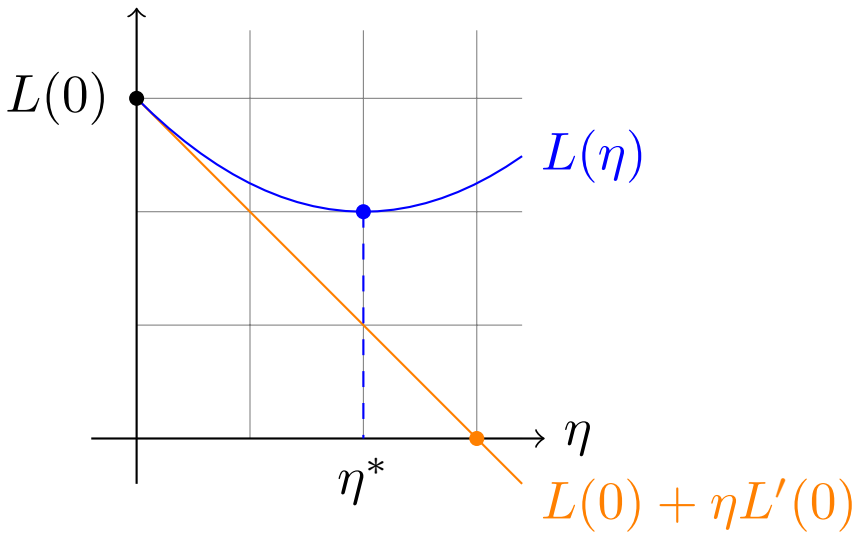

Unfortunately, the Stochastic Polyak stepsize is very unstable and can spike. 
This often happens in the beginning of training when $L(\theta_t)$ is large and the gradients are small (especially if the softmax is used). 
In the case of cross entropy loss, the curve does not reach zero and is locally convex for most minibatches. 
Hence, the Polyak stepsize overestimates, as illustrated in figures~\ref{fig:PolyakOverestimates} and~\ref{fig:comparison}.   

For this reason, many variants for practical use have been proposed~\cite{Berrada2020ALIGtraining,Loizou2021stochastic,King2024usingSPSinConvNetwork,Oikonomou2024SPSvariants,Orvieto2024GuassNewton,Orabona2025generalConvergence}.  However, they lead often lead to unstable training, especially when combined with momentum. Moreover, they are not able to replace manual learning rate selection because of sensitive dependencies on newly introduced hyper parameters. (Also for LLMs there are such reports~\cite{Orabona2026new}.) Perhaps, an exception is the {\em momo} scheduler from~\cite{Schaipp2024momo}, which is designed to optimize the loss aggregated from the last few minibatches. It works well with momentum and weight decay. Moreover, it has has only a single hyper parameter that is easy to tune, but it has not been tested systematically against competitors.

\subsection*{Second order methods} \label{ss:2ndorder}

\textbf{Approaches with test points.}
To select the learning rate, one may consider a quadratic approximation of the loss, 
\[
  \L(\theta_t -w) \approx \L(\theta_t) - w \cdot \nabla \L(\theta_t) + \tfrac 1 2 w^\mathsf{T} \Hess_{\L} w. 
\]
Each optimization method in a typical machinelearning optimizer defines a direction, 
for example, for basic SGD the direction is $w = -\eta \nabla L(\theta_t)$ or with momentum, it is $w = -\eta \nabla L(\theta_t) - \mu v_{t-1}$. 
After substituting in~$L$, one obtains an equation of the form $f(\eta) = a \eta^2 + b \eta + c$ from which an optimal stepsize $\eta$ can be computed. 
In~\cite{Bu2025gen,Zhu2021automatic}, the coefficients $a,b,c$ are obtained from 3 evaluations of~$f$. 
Since gradient descent already provides 1 evaluation, this method makes 2 extra forward evaluations, and hence, is $\%66$ slower (assuming a backwards pass takes twice the time of a forward pass). 
 These methods reduce the training loss very fast, but at the expense of the test loss~\cite{Zhu2021automatic}.
A similar technique is applied in~\cite{Bu2025gen} in which the computed learning rate is smoothed using a moving average and improvements on some fine tuning tasks are reported.\footnote{ 
  The paper~\cite{Iyer2021lrtuner} describes a similar algorithm with good performance, but its description is vague and no code is provided.
  }

In~\cite{Fu2024qlabgrad} it is noted that $w \cdot \nabla L(\theta_t)$ is fast to compute, thus to find the coefficients $a,b,c$, 
only 1 extra forward evaluation is needed. 
The paper presents an algorithm that is essentially parameter free and provably converges to a stationary point for Lipschitz functions. 
Unfortunately, they do not report test performance.

\bigskip
\noindent
\textbf{Barzilai-Borwein stepsizes.}
Another approach is to calculate the solution using the available gradient information, thus one needs
\begin{equation}\label{eq:eta_secondorder}
  \eta_t = \frac {v \cdot \nabla L(\theta_t)} {v^{\mathsf T} \Hess_L v}. 
\end{equation}
In the Barzilai-Borwein method~\cite{Barzilai1988twoStepsizes} one assumes $v = \nabla L(\theta_t)$ and estimates the above value by 
\[
  \eta_t = \frac {||\theta_t -\theta_{t-1}||^2} {(\theta_t - \theta_{t-1}) \cdot (\nabla L(\theta_t) - \nabla L(\theta_{t-1}))}. 
\]
This inspired a number of automated learning rate methods~\cite{Castera2022second,Tan2016bbsgd,Wang2023clippedBB}. 
Again, the method suffers from instability, but~\cite{Wang2023clippedBB} results come close to standard methods 
such as SGD with cosine annealing. 

\bigskip
\noindent
\textbf{1st order estimates of the 2nd order.}
In~\cite{Balboni2024adler}, the stepsize~\eqref{eq:eta_secondorder} is estimated directly with the already computed gradient. 
The denominator involves the Hessian, which is expensive to compute. 
However, typical networks of the last decade have wide hidden layers, and this allows for a good approximation with gradients. 
This is because preactivations of the output neurons are locally linear functions, as explained  
 in many papers about the neural tangent kernels, see for example~\cite{JacotNTK,BelkinLossLandscape,RobertsYaida2022book}.

For this, consider the loss as a function of output preactivations 
\[
  L(\theta) = \mathcal L(F(\theta)). 
\]
The Hessians of $L$ can be written using the Hessians of $\mathcal L$ and the Hessians of the output preactivations~$F_a$. 
Let $\Jac_{a\mu} = \frac {\partial F_a} {\partial \theta_\mu}$ be the Jacobian of~$F$. 
\[
  \Hess_L = \Jac \Hess_{\mathcal L} \Jac + \sum_a \frac {\partial \mathcal L} {\partial F_a} \Hess_{F_a}. 
\]
For large neural nets, the spectral norm of the Hessians $F_a$ are proportional to $O(\text{depth}/\text{width})$, see~\cite{RobertsYaida2022book}. 
In the {\em outer product approximation} of $\Hess_L$ all these Hessian are neglected and only the 1st term remains, see for example~\cite[p 243, sect.8.1.6]{Bishop2024deep}. 
In~\cite{Balboni2024adler} this first term is computed using the available gradient, and~\eqref{eq:eta_secondorder} is obtained with negligible computational overhead. 
They show in a few experiments that this reduces the train error better than SGD, but do not report test errors.  

\bigskip
\noindent
Below, we propose a method that uses a single test point to estimate this 2nd order stepsize in the same way.

\subsection*{Other methods}

In the paper~\cite{Liu2022leap}, a wide range of optimizers is modified in a generic way: 
in each minibatch, the existing learning rate schedule is randomly increased or decreased by a few percentages. 
Despite its simplicity, this method is reported to consistently increase test error, including on Cifar10 and Resnet18 experiments. 

This easy method is similar to other regularization methods like dropout~\cite{Srivastava2014dropout} or neftune~\cite{Jain2024neftune} that increase randomness during training.
The motivation is that solutions with wide minima have better generalization and it is argued that these perturbations can help to escape narrow local minima. 
See~\cite{Xie2020diffusion} for this theory.

Finally, the paper~\cite{Cutkosky2023mechanic} uses a complex algorithm from the field of online convex optimization to select the learning rate. 
[todo: test error?]

\section{Two proposed stepsizes}

We aim to minimize to optimize parameters $\theta$ to train a neural nets. 
For this, a standard run of stochastic gradient descend is applied, which means that we start from a randomly generated $\theta_0$ 
and at each step, a minibatch loss $\L_t(\theta)$ is used to update $\theta_t$ using
\begin{align}
  v_{t} & = \mu v_{t-1} + \eta_t \nabla \L_t(\theta_{t-1}+\mu v_{t-1}) + \lambda \theta_{t-1}, \label{eq:SutskeverUpdate} \\
   \theta_{t} & = \theta_{t-1} - v_t \nonumber, 
\end{align}
where $\mu$ and $\lambda$ are the parameters for momentum and weight decay. 

\noindent
\textbf{Notation.}
Consider a neural net with $\nout$ output neurons and a train set whose examples are denoted by $(x_\alpha, y_\alpha)$ with Greek indices. 
For each training example, the label $y_\alpha$ is given as a one-hot encoding of the correct label, thus $y_{\alpha c} = 1$ if $x_\alpha$ belongs to the $c$-th class. 
The preactivations of the output neurons are denoted by the vector $f(x_\alpha,\theta)$, where $\theta$ is the list of all trainable weights of the net. 

\bigskip
\noindent
We now define the minibatch loss $\L_t$.  Let
\begin{align*}
  z_{\alpha i}            &= f_i(x_\alpha, \theta) \\
  q_{\alpha i}            &= \frac {\exp z_{\alpha  i}} {\sum_{j \le \nout} \exp z_{\alpha j}} \\
\end{align*}
For a list of train examples~$S_t$, the $t$-th {\em minibatch loss} is
\[
  \L_t(\theta)= - \sum_{\alpha \in S_t} \sum_{i \le \nout\!\!\!\!} y_{\alpha i} \ln q_{\alpha i}. 
\]

\noindent
\textbf{Notation.}
We write $\sum_\alpha$ to denote $\sum_{\alpha \in S_t}$ and sums over indices $i,j,k$ range from $1$ to~$\nout$.

\subsection*{The second order outer approximated stepsize (so-dz)}

Fix $t$ and let $L(\tilde\eta) = \L_t(\theta_t - \tilde\eta v_t)$. 
We select the stepsize $\eta_{\text{so}}$ to minimize the quadratic approximation of $\L_t(\tilde \eta)$. 
As explained in subsection~\ref{ss:2ndorder}, we may assume that $z_{\alpha i}$ is linear in $\tilde \eta$ and we can obtain the derivative from only the gradient. 
Still, to increase stability, we use a test point. 

\medskip
\noindent
Let us explain this plan in more details.  We have
\[
  L_t(\eta) \approx L_t(0) - \eta L'(0) + \frac {\eta^2} {2} L''(0). 
\]
To compute $L'(0)$ we view $\L$ as a function of the output preactivations $z_{\alpha i}$ and apply the chain rule. 
The 2nd derivative $L''(0)$ is computed similarly, while the 2nd derivatives of $z_{\alpha i}$ are neglected, see appendix~\ref{ss:sodz-derivation} for details. 
Let $\delta_{ij}$ be the Kronecker function, which equals $1$ if $i=j$ and $0$ otherwise. 

  \[
  \eta_{\text{so}}^* = \frac {L'(0)} {L''(0)} = \frac 
            {\sum_{\alpha, i} (y_{\alpha i} - q_{\alpha i}) \tfrac {\dd z_{\alpha i}} {\dd \eta}}
	    {\sum_{\alpha, i, j} (\delta_{ij} - q_{\alpha i})q_{\alpha j} \tfrac {\dd z_{\alpha i}} {\dd \eta} \tfrac {\dd z_{\alpha j}} {\dd \eta}}.
	  \]
Finally, in this expression, $\frac {\dd z_{\alpha j}} {\dd \eta}$ is estimated by evaluating the output preactivations $z$ with a test update of $\eta$ 
using a fixed learning rate $\eta^{\tst}$. 
More precisely, let 
\[
  \theta_t^\tst = \theta_t - \mu v_t - \eta^\tst \nabla \L(\theta_t) - \lambda \theta_t. 
\]
Let $q^\tst_\alpha$ and $q_\alpha$ be the softmax values evaluated with parameters $\theta_t^\tst$ and $\theta_t$. 
Then, estimate $\frac {\dd z_{\alpha}} {\dd \eta}$ by $(z^\tst_\alpha - z_\alpha)/\eta^\tst$. Thus, 
\begin{equation}\label{eq:eta_so}
  \etaso = \eta^\tst \frac 
	    {\sum_{\alpha, i} (y_{\alpha i} - q_{\alpha i}) (z^\tst_{\alpha i} - z_{\alpha i})}
	    {\sum_{\alpha, i, j} (\delta_{ij} - q_{\alpha i})q_{\alpha j}  (z_{\alpha i}^\tst - z_{\alpha i})(z_{\alpha j}^\tst - z_{\alpha j})}. 
\end{equation}
This finishes the construction. The pseudocode is given in algorithm~\ref{alg:NetLineCrossEntropy}. 
In the experiments, we fixed $\eta^\tst = 5 \cdot 10^{-5}$. 

\begin{algorithm}[t]
  \caption{Stepsize modulation algorithms with $\etaso$ and $\etacs$}
  \label{alg:NetLineCrossEntropy}

  \begin{algorithmic}
    \renewcommand{\L}{\mathcal L}

    \State \textbf{Initialize:} $\theta_0$ randomly, $v_0 = 0$

    \For{$\mathrm{epoch} = 1, \ldots, T$}

	\State sample a fixed-size mini-batch $S$ and
	use data $\{(x_\alpha, y_\alpha)\}_{\alpha \in S}$

	\State $\begin{array}[t]{@{}l@{\;\gets\;}l@{}}
	    \eta_{\mathrm{epoch}}
	    & \operatorname{schedulerValue}(\mathrm{epoch}) \\
	    \vartheta_t
	    & \theta_t - (\mu v_t + \lambda \theta_t) \\
	    z_\alpha
	    & f(x_\alpha, \vartheta_t) \\
	    q_\alpha
	    & \operatorname{softmax}(z_\alpha) \\
	    \L_t
	    & \operatorname{crossEntropy}(z_S, y_S) \\
	    \theta^{\tst}
	    & \vartheta_t - \eta^{\tst} \nabla_{\theta} \L_t \\
	    z^{\tst}_\alpha
	    & f(x_\alpha, \theta^{\tst}) \\
	    q^{\tst}_\alpha
	    & \operatorname{softmax}(z^{\tst}_\alpha) \\
	    \hat{\eta}
	    & \text{eq~\eqref{eq:eta_so} or \eqref{eq:eta_cs} using } z_S, z^{\tst}_S, q_S, q^{\tst}_S \\
	    \eta_{t+1}
	    & \operatorname{modulation}
	    (\hat{\eta}, \eta_{\mathrm{epoch}}) \\
	    \theta_{t+1}
	    & \theta^{\tst}
	    - (\eta_{t+1} - \eta^{\tst}) \nabla_{\theta} \L_t \\
	    v_{t+1}
	    & \mu v_t + \lambda \theta_t
	    + \eta_{t+1} \nabla_{\theta} \L_t
	\end{array}$

    \EndFor

    \State \Return $\theta_{T+1}$

    \end{algorithmic}

\end{algorithm}

\subsection*{The closest softmax stepsize (cs-dq)}

A linear approximation of the softmax values in $\eta$ is used to select a value that minimizes the average Euclidean distance to the one-hot encodings $y_{\alpha}$. 
Thus we minimize the sum of their $2$-norms
\[
  \tfrac 1 2 \sum_\alpha \bignorm{ y_\alpha - q_\alpha - \eta \dqdeta}^2. 
\]
Recall that sums over $\alpha$ range over the $t$-th minibatch $S_t$.
Solving the derivative to $\eta$ gives
\[
  \eta_{\text{cs}}^* = \frac {\sum_\alpha (y_\alpha - q_\alpha) \cdot \dqdeta} {\sum_\alpha \big\lVert \dqdeta \big\rVert^2}.  
\]
Finally, we use the same probe for $\eta^\tst$ as before to estimate~$\dqdeta$. 
\begin{equation}\label{eq:eta_cs}
  \etaso = \eta^\tst \frac 
	    {\sum_\alpha (y_\alpha - q_\alpha) \cdot (q^\tst_\alpha - q_\alpha)} 
	    {\sum_\alpha \big\lVert q^\tst - q_\alpha \big\rVert^2}.  
\end{equation}

\subsection*{Both stepsizes are close}

\begin{figure}
  \centering
    \includegraphics[width=0.4\textwidth]{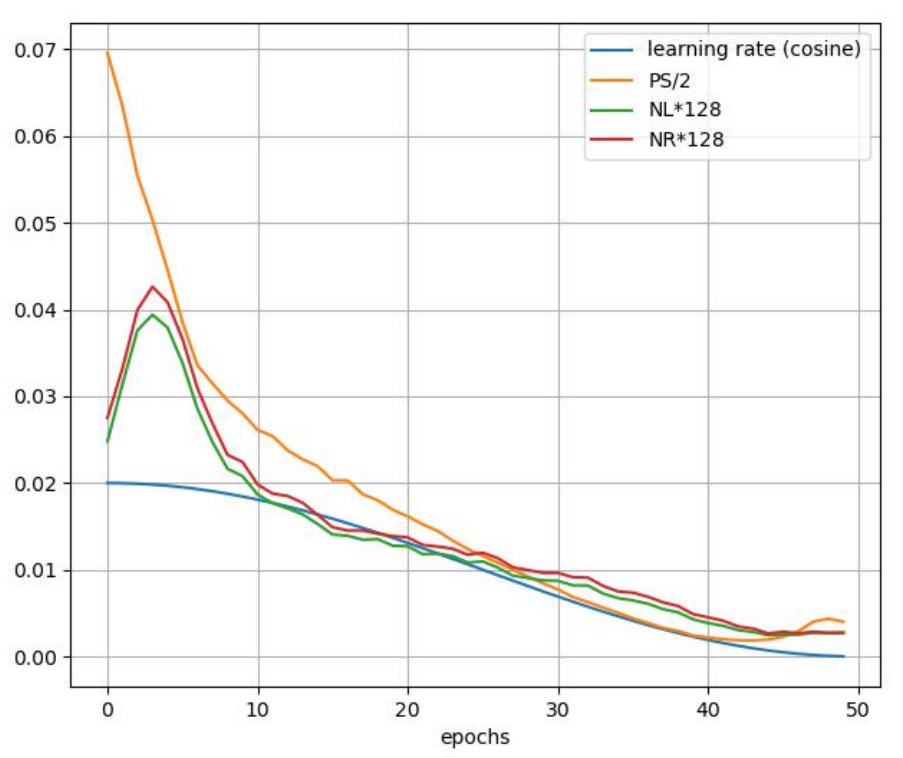}
  \caption{comparison between stepsizes: so-dz (red),}

  cs-dq (green), Polyak (orange), cosine decay (blue). 
  \label{fig:comparison}
\end{figure}

Despite being derived by different philosophies, the averages of these learning rates are close to each other. 
Indeed, we trained Resnet18 on Cifar10 with standard SGD (max learning rate $0.02$, momentum $\mu = 0.9$, weight decay $\lambda = 0.005$) 
and cosine learning rate schedule. During this run we computed the average values of the Polyak stepsize and the 2 above stepsizes. 
In figure~\ref{fig:comparison}, we compare these stepsizes. 
We conclude that after halving the Polyak stepsize, these values are close to each other during the middle part of the training. 

\bigskip
\noindent
[To do: discuss fluctuations and compare values during minibatches.]

\section{Practical refinements}

\subsection*{Outlier removals in stepsize estimates}

\begin{figure}
  \centering
  \quad\includegraphics[width=0.42\textwidth]{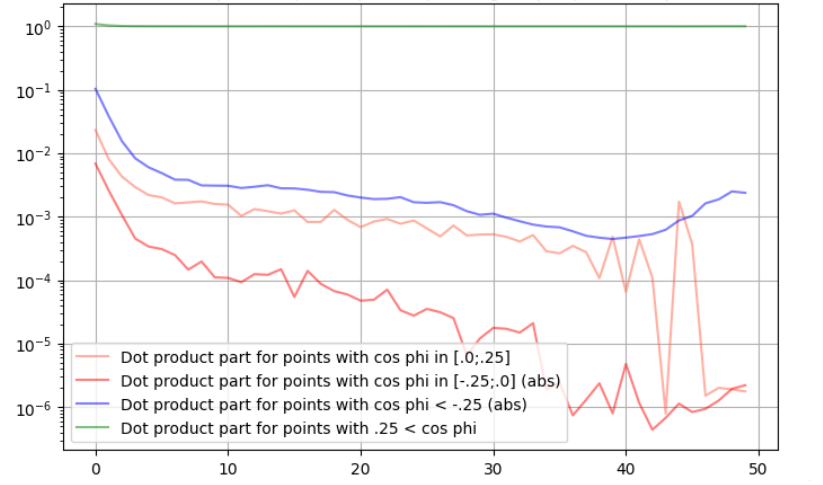}
  \caption{Fraction of removed outliers in the stepsize calculation}
  \label{fig:outlier_removal}
\end{figure}

If for some training example $\alpha$ of a minibatch, the softmax values do not seem to improve, then we drop this input in the estimation of $\etaso$ and $\etacs$. 
More precisely, we drop $\alpha$ from minibatch $S_t$ if $\cos \phi_\alpha < 1/4$ where
\[
  \cos \phi_\alpha = \frac {(y_\alpha - q_\alpha) \cdot (q^\tst_\alpha - q_\alpha)} {||y_\alpha - q_\alpha|| \,||q^\tst_\alpha - q_\alpha||}.
\]
The effect is small, because only few points satisfy the condition $\cos(\phi_\alpha) < 1/4$, as shown in figure~\ref{fig:outlier_removal}. 
But it seems to help to stabilize the estimates [can we say something more convincing or precise]?

\subsection*{Subsampling and computational overhead}

The estimation of the stepsize requires an extra forward pass. 
This increases the runtime typically by~$33\%$, because the time for a backward pass is usually twice the time of a forward pass. 

In our 2nd series of experiments, we took 20 random inputs from the remaining minibatch $S_t$, (which had 128 elements, before the above pruning),
and estimated the learning rate from them. 
This reduced the overhead from $33\%$ to about~$21\%$ (or better on GPU's with few cores and fast memory).

\subsection*{Modulate an existing learning rate schedule}

The effect of changing the learning rate for a given step also has an effect on future steps, because of momentum. 
Several papers report that this is a challenge to design automated learning rate schemes, for example in~\cite[footnote 1 on p7]{Vaswani2019painless}. 
Moreover, the several theoretical studies suggest that during initial learning it is beneficial to keep the stepsize larger, 
in order to avoid parameters $\theta_t$ approaching sharp local minima. 
For these reasons, we focussed on perturbing existing schedules, rather than reconstructing them from scratch. 

\begin{itemize}
  \item 
    Let $\eta_{\text{ref}, e}$ be a reference learning schedule evaluated in epoch~$e$. 
    In experiments we will use cosine decay. 
  \item 
    Let $\eta_{t}$ be the stepsize from 1 of the above methods, thus $\etaso$ or $\etacs$ at batch~$t$. 
  \item 
    Let $\bar \eta_t$ be the average of $\eta_t$ over last $N = 50$ batches. 
  \item 
    Let $\beta = 0.05$ be fixed. [Todo: it turns out that we have only lookahead during experiments of our paper.]
\end{itemize}
Rather than clipping, a smoothly perturbed version of $\eta_{\text{ref}, e}$ is obtained by multiplyin it with a the following factor in $[1-\beta \pi/2, 1+\beta \pi/2]$, 
\[
  \eta_{\text{final},t} = \eta_{\text{ref},e} \Big( 1 + \beta \arctan \frac {\eta_t - \bar \eta_t} {\beta \bar\eta_t} \Big).
\]
The $\arctan$ function was used because the derivative in the origin equals~1.

\subsection*{Lookahead}

In the 1st series of experiments, the lookahead technique from~\cite{Zhang2019lookahead} was applied. 
The idea is to stabilize an existing optimizer Opt (like SGD, AdamW, etc) by letting it first make $k$ updates and then  
make a single update of the aggregated ``gradients''. 
The optimizer has 2 parameters $k$ and $\alpha$. We use the default values $k=5$ and~$\alpha = 0.8$. 

More precisely, the optimizer Opt is viewed as a fast optimizer and is used to create a slow optimizer as follows. 
Let $\theta_t$ denote the current parameters of the slow optimizer at the start of its slow step. 
The fast optimizer is initialized with $\phi_{t,0} = \theta_t$ and computes $k$ updates $\phi_{t,1}, \ldots, \phi_{t,k}$. 
Finally, the slow optimizers sets 
\[
  \theta_{t+1} = (1-\alpha) \theta_t + \alpha \phi_{t,k}.
\]

\section{Experiments}

We use the well known Resnet18 and Resnet34 models and trained them from scratch on Cifar10 and Cifar100. 
Data augmentation and the modification of the last hidden layer was applied as is now standard, see for example~\cite{He2019bag}.

The 2 baseline experiments are as follows. 
\begin{itemize}
  \item 
    Standard SGD with cosine learning rate decay (without warmup). 
    The max-learning rate is $0.02$, the momentum parameter is $\mu = 0.9$, the weight decay is $5e-3$ and the minibatch size is~128. 
  \item 
    Lookahead with the above SGD as fast optimizer. 
    The standard values $k = 5$ and $\alpha = 0.8$ are used. 
\end{itemize}
We have a default experiment which uses Resnet18 on Cifar10 with 50 epochs. 
All pictures where generated with this default experiment.  
The max-learning rate and weight decay parameters reported above where tuned for this experiment. 
Our 2 stepsizes were designed using this same experiment. 
However, below we report these baselines for different settings with 100 and 200 epochs, and we also report the experiments with Resnet34 and Cifar100. 
In all experiments the hyper parameters remain the same and are fixed. 

\begin{figure}[h]
    \centering
    \includegraphics[width=\textwidth]{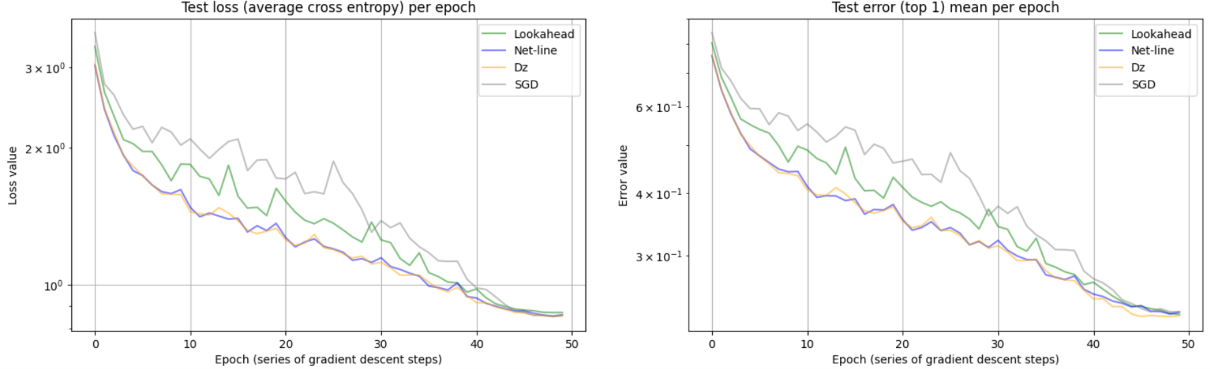}
    \caption{Loss and test error per epoch. Experiment with CIFAR100 and ResNet18, 50 epochs, lookahead.}
    \label{fig:loss_error_cifar100_rn18}
\end{figure}

\bigskip
\noindent
In all experiments, the hyper parameters of the stepsize $\etaso$ and $\etacs$ are $\eta^\tst = 5 \cdot 10^{-5}$, the perturbation $\beta = 0.05$ and the average stepsize is computed from $50$ minibatches. Experiments were performed on V100 GPU's on the cluster of the Higher School of Economics in Moscow~\cite{Kostenetskiy2026HPC}. The code is available here~\url{https://github.com/Brbauwens/adaptive_stepsize_modular}. 

For both the so-dz and cs-dq stepsizes, the initial test error decreases more rapidly and the curve is more monotone. 
However, recall that this method requires $33\%$ more compute (and $21\%$ more compute with subsampling of the test probe). 
Especially, this is true when combined with lookahead, see figure~\ref{fig:loss_error_cifar100_rn18} for the default experiment.  
However, near the end, improvement becomes slower.

Our methods often improve error on the test set. 
In a 1st series of experiments, lookahead was applied and no subsampling was done. 
All 4 methods were applied to the same minibatches. 
The best test set accuracies are given in tables~\ref{tab:experiments1}. 
For the Cifar10 training set, both the so-dz and cs-dq stepsizes exceed both baselines. 
Unfortunately, for Cifar100 the results look random. 

In a 2nd series of experiments, subsampling was applied and lookahead was switched off. 
The runs of the baselines were repeated with different random seeds. 
The results in table~\ref{tab:experiments2} suggest that this does not decrease performance. 
Again for Cifar10, the best results is either for so-dz or cs-dq, but in the case of Resnet34, the gains look random. 
Also, for Cifar100, the methods do not provide reliable improvements.

\newcommand{\colw}{2.2cm}
\newcommand{\mb}{\mathbf}

\begin{table}
  \centering
    \begin{tabular}{ |p{1.3cm}|p{1.8cm}|p{\colw}|p{\colw}|p{\colw}|p{\colw}|  }
     \hline
        &  &\multicolumn{2}{c}{Cifar10}\vline & \multicolumn{2}{c}{Cifar100}\vline \\
     \hline
        Net & optimiser &100 epochs& 200 epochs&100 epochs&200 epochs    \\
     \hline
  Resnet18 & SGD          & $95.256\pm0.152$      & $95.476\pm0.099$      & $78.063\pm0.239$      & $\mb{78.649}\pm0.259$\\
           & la           & $95.287\pm0.12$       & $95.603\pm0.149$      & $78.123\pm0.296$      & $78.543\pm0.254$\\
           & la, so-dz    & $\mb{95.399}\pm0.127$ & $95.608\pm0.102$      & $\mb{78.157}\pm0.229$ & $78.593\pm0.241$\\
           & la, cs-dq    & $95.379\pm0.097$      & $\mb{95.673}\pm0.167$ & $78.114\pm0.277$      & $78.554\pm0.204$\\
     \hline
  Resnet34 &  SGD         & $95.342\pm0.106$      & $95.536\pm0.127$      & $78.891\pm0.249$        & $79.357\pm0.259$\\
           &  la          & $95.471\pm0.134$      & $95.629\pm0.153$      & $\mb{78.968}\pm0.265$   & $\mb{79.537}\pm0.197$\\
           &  la, so-dz   & $95.550\pm0.109$      & $\mb{95.711}\pm0.097$ & $78.919\pm0.225$        & $79.461\pm0.174$\\
           &  la, cs-dq   & $\mb{95.574}\pm0.102$ & $95.663\pm0.087$      & $78.959\pm0.204$        & $79.519\pm0.256$\\
     \hline
    \end{tabular}
    \caption{Test accuracy, averaged over 16 repetitions}
    \label{tab:experiments1}

    \bigskip

    \begin{tabular}{ |p{1.3cm}|p{1.8cm}|p{\colw}|p{\colw}|p{\colw}|p{\colw}|  }
     \hline
        &  &\multicolumn{2}{c}{Cifar10}\vline & \multicolumn{2}{c}{Cifar100}\vline \\
     \hline
        Net & Optimiser &100 epochs& 200 epochs&100 epochs&200 epochs    \\
     \hline
 Resnet18   & SGD        & $95.266      \pm 0.164$  & $95.552 \pm 0.130$      & $77.896      \pm 0.262$ & $78.610 \pm 0.404$\\
	    & lookahead  & $95.294      \pm 0.148$  & $95.520 \pm 0.154$      & $\mb{78.225} \pm 0.263$ & $\mb{78.723} \pm 0.179$\\
	    & la, so-dz & $95.363      \pm 0.071$  & $95.558 \pm 0.147$      & $78.176      \pm 0.153$ & $78.595 \pm 0.180$\\
            & SGD, cs-dq & $\mb{95.425} \pm 0.134$  & $\mb{95.653} \pm 0.135$ & $78.051      \pm 0.212$ & $78.353 \pm 0.300 $\\
     \hline
 Resnet34 & SGD         & $95.436      \pm 0.132$  & $ 95.428 \pm 0.198$      & $78.833 \pm 0.249$      & $ 79.347 \pm 0.190 $ \\
	  & lookahead   & $95.438      \pm 0.165$  & $ 95.667 \pm 0.139$      & $78.863 \pm 0.257$      & $ \mb{79.592} \pm 0.257 $ \\
          & SGD, so-dz  & $\mb{95.574} \pm 0.153$  & $ 95.625 \pm 0.151$      & $\mb{78.941} \pm 0.236$ & $ 79.415 \pm 0.141 $ \\
	  & SGD, cs-dq  & $95.433      \pm 0.107$  & $ \mb{95.703} \pm 0.158$ & $78.876 \pm 0.302$      & $ 79.427 \pm 0.289 $ \\
     \hline
    \end{tabular}
    \caption{Test accuracy averaged over 6 repetitions, subsampling}
    \label{tab:experiments2}
\end{table}

\section{Conclusion}

Two methods were proposed to automatically generate a stepsize for each minibatch. 
In $\etaso$ stepsize optimizes a 2nd order approximation of the minibatch loss, 
and the $\etacs$ optimizes the softmax values. 
We argued theoretical and experimentally that both methods give closely related stepsizes. 

These stepsizes were used to improve the cosine decay schedule. 
Experiments show that they speed up convergence beyond the 20\%-33\% especially in the middle of the training. 
Moreover, the decrease is more monotone. 
They give essentially the same accuracy, and in a few experiments, their best accuracy is improved compared to standard SGD and its combination with lookahead.

\bibliographystyle{plain}
\bibliography{references}

\appendix

\section{Derivation of so-dz stepsize}\label{ss:sodz-derivation}

Fix the minibatch is fixed and we drop subscripts~$t$. 
Below, the quantities $q$ and $z$ are considered for the updated parameter $\theta_{t+1}$, which 
depends on the updated learning rate through \eqref{eq:SutskeverUpdate}, 
thus $q = q(\theta_{t+1}(\eta))$ and $z = z(\theta_{t+1}(\eta))$. 
\begin{align*}
  q_{\alpha i}   &= \frac {\exp z_{\alpha  i}} {\sum_{j} \exp z_{\alpha j}} \\
  L(\tilde\eta)  & = \L(\theta - \tilde\eta w). 
\end{align*}
We need to evaluate $L'(0)$ and $L''(0)$ and write it as a function of $q$ and $\frac {\dd z} {\dd \eta}$. 
Recall that $y_{\alpha j}$ is the one hot encoding of the label of training input~$x_\alpha$. 
One calculates that
\begin{align}
  \frac {\partial q_{i\alpha}} {\partial z_{j\alpha}} & = (\delta_{ij} - q_{\alpha i}) q_{\alpha j} \label{eq:dqdz}\\
  \frac {\partial L} {\partial z_{i\alpha}} & = q_{\alpha i} - y_{\alpha i} \nonumber\\
  \frac {\partial^2 L} {\partial z_{i\alpha} \partial z_{j\alpha}} & = (\delta_{ij} - q_{\alpha i}) q_{\alpha j}. \nonumber
\end{align}
We have 
\[
  L'(0)   = \sum_{\alpha,i}   \frac{ \partial L} {\partial z_{\alpha i}} \dzideta  = \sum_{\alpha, i} (q_{\alpha i} - y_{\alpha i}) \dzideta. \\
\]
To calculate $L''$, we neglect the 2nd derivatives of the output preactivation~$z$. Thus, 
\begin{align*}
  L''(0)  = \sum_{\alpha,j} \frac{ \partial L'} {\partial z_{\alpha j}} \dzjdeta  
  & = \sum_{\alpha, i, j} \frac {\partial^2 L} {\partial z_{j\alpha} \partial z_{i\alpha}} \dzideta \dzjdeta \\
  & = \sum_{\alpha, i, j} (\delta_{ij} - q_{\alpha i}) q_{\alpha j} \dzideta \dzjdeta. 
\end{align*}

%
%

\end{document}